\documentclass[10pt,a4paper]{article}
\usepackage[utf8]{inputenc}
\usepackage[T1]{fontenc}
\usepackage[margin=2.5cm]{geometry}
\usepackage{lmodern}
\usepackage{setspace}
\usepackage{graphicx}
\usepackage{booktabs}
\usepackage{amsmath}
\usepackage{url}
\usepackage[hidelinks]{hyperref}
\usepackage[round,authoryear]{natbib}
\usepackage{subcaption}
\usepackage{placeins}

\title{From Nowcasting to Forecasting: Adapting a Reanalysis-Trained Cloud Cover Model to Observations}

\author{
Mikko Partio$^{1,2}$\thanks{Corresponding author: mikko.partio@fmi.fi}
\and
Leila Hieta$^{1}$
\and
Ossi Laine$^{1}$
}

\date{
$^{1}$Finnish Meteorological Institute\\
$^{2}$Aalto University\\[1em]
\today
}

\begin{document}

\maketitle

\begin{abstract}
Accurate cloud-cover forecasts are important for temperature prediction, radiation forecasting, and solar-power operations. Short-range forecasting methods can preserve observed cloud placement during the first forecast hours, but their skill decreases when cloud fields evolve through formation, dissipation and deformation. Longer lead times require accounting for atmospheric evolution, but operational numerical weather prediction (NWP) forecasts may not accurately represent the satellite-observed cloud state at initialization. We develop CloudCast v2, a machine-learning model for 12-hour cloud-cover forecasting from observation-based initial conditions. The model is first trained on the Copernicus European Regional Reanalysis \citep{Ridal2024} to learn cloud-evolution dynamics, and is then adapted to satellite-derived cloud fields using conditional flow matching \citep{Lipman2023}, a generative method that transforms noise into cloud-cover forecasts conditioned on the observed initial cloud fields and NWP inputs. CloudCast v2 reduces mean absolute error by 10\% relative to its predecessor, CloudCast v1 \citep{Partio2025}, over the 1--12 h range. It also overtakes CloudCast v1 in fractions skill score, a neighborhood-based measure of spatial agreement, after approximately 3--6 h, depending on the cloudiness category. These results show that observation-initialized machine-learning forecasts can extend beyond the usual 1--3-hour nowcasting range while retaining spatial detail from satellite cloud fields.
\end{abstract}


\section{Introduction}

% GENERAL INTRODUCTION

Cloud cover is a major regulator of the Earth’s radiation budget and an important driver of both weather and climate \citep{Hughes2024}. Accurate cloud-cover forecasts are therefore valuable not only for meteorological prediction, but also for applications such as solar-power management and energy-system operations. Errors in cloud prediction can propagate into substantial biases in near-surface temperature, radiation, and other forecast variables. At local scales, the errors increase uncertainty in solar energy generation and decrease the reliability of operational decision-making \citep{Wang2022}.

% NARROWING DOWN THE ISSUE

Short-range cloud forecasting sits between observation-based nowcasting and dynamical weather prediction. Across the 1--12 hour range considered here, forecasts commonly rely on either numerical weather prediction (NWP) or observation-based extrapolation. NWP models can represent atmospheric dynamics and are therefore able to predict evolving cloud structures beyond simple advection. Cloud cover remains difficult to predict because it depends strongly on moisture and vertical motion, as well as on subgrid-scale processes that must be parameterized. Moreover, initial-state errors can strongly affect forecast quality \citep{Sun2014}. Extrapolation methods, in contrast, start from the latest observations and estimate future cloud fields by moving the observed cloud patterns according to their recent motion. These methods are often effective at lead times of one to two hours, when much of the forecast skill comes from persistence and advection \citep{Prudden2020}. At longer lead times of three hours or more, their skill drops as cloud formation and dissipation become important \citep{Partio2025}.

% PRESENTING THE PROBLEM

In operational forecasting, users need a single forecast covering the full lead-time range rather than separate forecasts for different lead times. This requirement creates a difficult transition: extrapolation-based methods tend to perform best at the earliest lead times, while dynamical models become more useful as lead time increases. Previous work has blended the approaches in image space across different lead-time ranges \citep{Hieta2021}; however, producing one coherent forecast remains nontrivial and can introduce visible discontinuities when the underlying forecasts disagree strongly. This transition between nowcasting and forecasting remains a central challenge for operational cloud-cover prediction.

Research on observation-based short-range forecasting has traditionally focused on extrapolating observed fields, particularly radar-derived precipitation fields, while more recent machine-learning approaches learn their evolution from data \citep{Prudden2020}. Recurrent neural networks (RNNs) were among the early deep-learning approaches \citep{Shi2015} and were followed by convolutional architectures \citep{Ayzel2020,Ha2024,Ritvanen2023}. While deep generative radar nowcasting has improved precipitation forecasts at lead times of 5--90 minutes \citep{Ravuri2021}, other machine-learning models have extended high-resolution precipitation forecasting to 12 hours \citep{Espeholt2022}. More recently, conditional flow matching, a generative method that transforms random noise into forecast fields conditioned on the input data, has been applied to probabilistic precipitation nowcasting \citep{Ribeiro2026}. These studies have, however, mainly addressed radar-observed precipitation. Satellite-derived cloud fields represent a different prediction target, so the applicability of these methods to cloud-cover forecasting remains uncertain.

Cloud-cover forecasting has received less attention than precipitation nowcasting. It poses a related but distinct prediction problem: cloud cover is a bounded fractional field, with predominantly clear-sky and overcast regimes and complex intermediate structures associated with partial or broken cloud cover. Recent studies have begun to target cloud cover directly with dedicated machine-learning models. For example, \citet{Kellerhals2022} and \citet{Xia2024} developed RNNs to predict cloud cover, while \citet{Partio2025} introduced CloudCast v1, a U-Net-based model trained on satellite-derived effective cloudiness to forecast cloud cover up to five hours ahead, with a reported skillful lead time of about three hours. Generative approaches have also been applied to satellite imagery. \citet{Chase2026} investigated score-based diffusion for GOES-16 nowcasting, while \citet{Chen2026} introduced SATcast, a cascade diffusion model conditioned on atmospheric-model fields. Both methods predict brightness temperature rather than fractional cloud cover.

In parallel, machine-learning weather forecasting has shown that models trained on reanalysis data can learn to predict atmospheric evolution beyond the nowcasting range. At synoptic and global scales, this progress includes GraphCast \citep{Lam2023}, PanguWeather \citep{Bi2022}, and AIFS \citep{Lang2024}. Regional studies have extended the same idea to limited-area forecasting, including graph-based neural weather prediction \citep{Oskarsson2023} and stretched-grid architectures \citep{Nipen2024}. These models target forecast lead times of several days, but they do not yet show how reanalysis-trained forecasting systems can be adapted to observation-based initial conditions for short-range prediction. Recent work has begun to connect these approaches by combining observations with NWP inputs in regional weather forecasting \citep{Miralles2026}, but adapting a reanalysis-trained model to satellite-derived fractional cloud cover for 12-hour forecasting remains underexplored.

% These developments show that the two strands of research are beginning to converge. Observation-based nowcasting methods typically start from observations and focus on short lead times, whereas machine-learning weather forecasting models learn from simulation or reanalysis and target longer forecast horizons. Other recent work also connects these areas \citep{Miralles2026}. However, adapting a reanalysis-trained forecasting model to satellite-derived fractional cloud cover while retaining skill across a 12-hour forecast horizon remains an open challenge.

% OUR SOLUTION

We therefore develop CloudCast v2, a unified machine-learning approach for 12-hour cloud-cover forecasting. First, we train a vision transformer-based model on reanalysis data to learn to predict cloud evolution on a 5~km grid covering Scandinavia. We then adapt the model to observation-based cloud fields using a clean-target formulation based on conditional flow matching \citep{Lipman2023}. The model also receives atmospheric fields from NWP forecasts at each forecast lead. This design allows the model to retain dynamics learned from reanalysis while aligning its forecasts with observed cloud fields. We evaluate CloudCast v2 against nowcasting and NWP baselines using pixel-wise error, neighborhood-based spatial verification, and skill scores for cases with large cloud-cover changes. We show that CloudCast v2 can be initialized from observed cloud fields while extending useful forecast skill to longer lead times than CloudCast v1.

% HANDOFF

The remainder of this paper is organized as follows. Section 2 describes the data and methods, including the model architecture and training setup. Section 3 presents the forecast evaluation results, and Section 4 discusses their implications and outlines directions for future research.

\section{Data and methods}

This section defines the data, forecasting task, model, training protocol, and evaluation framework used in the study. We first describe the regional domain and the three data sources: the Copernicus European Regional Reanalysis (CERRA) \citep{Ridal2024} for the long reanalysis record, Nowcasting Satellite Application Facilities (NWCSAF) effective cloudiness \citep{Kerdraon2021} as the observation-space target, and MetCoOp Ensemble Prediction System (MEPS) forecasts \citep{Eresmaa2026}, whose atmospheric fields provide time-varying inputs to CloudCast v2 and whose cloud-cover forecasts serve as the NWP baseline. We then describe the architecture of CloudCast v2, how the model is trained and adapted to observations, and how its forecasts are evaluated.

\subsection{Study domain and datasets}

% 1. Define the study domain
% State the spatial projection, grid spacing, geographical coverage, and domain figure.

\begin{figure}[!htbp]
\centering
\includegraphics[width=1.0\textwidth]{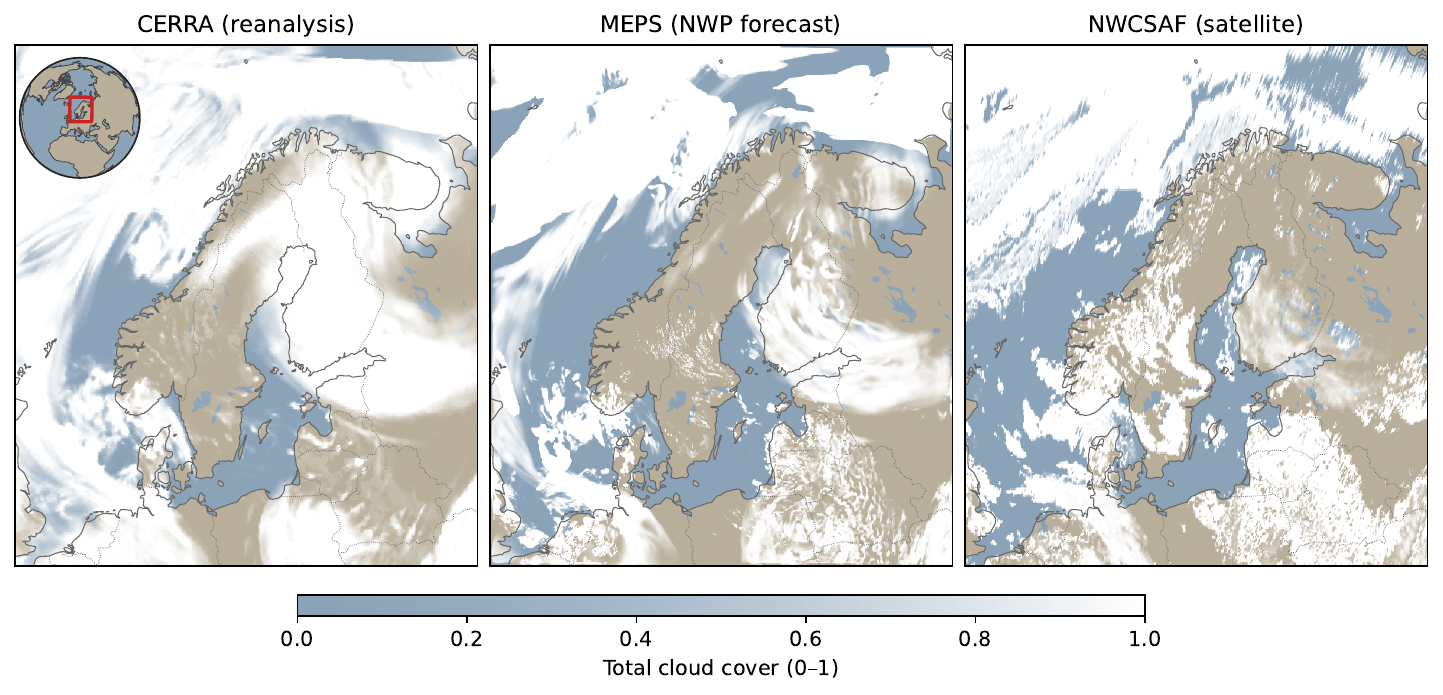}
\caption{Comparison of cloud-cover fields on the common 5~km grid for 3 June 2021 at 12 UTC. The panels show CERRA and MEPS total cloud cover, and NWCSAF effective cloudiness, for the same valid time. The persistent feature in the northeastern corner of the NWCSAF panel is a known retrieval artefact in the satellite data and is retained in the data used here.}
\label{fig:tcc_comparison}
\end{figure}

We conduct the study on a regional domain covering Scandinavia, defined on a common 5~km Lambert conformal conic grid. Before training and evaluation, we reproject all cloud-cover and atmospheric fields from their native grids to this common grid so that the different data sources are spatially aligned. Figure \ref{fig:tcc_comparison} compares the CERRA, MEPS, and NWCSAF cloud fields on this grid. In the example shown, all three exhibit the same broad cloud pattern, but CERRA and MEPS have less small-scale variability and fewer values near 0 and 1 than NWCSAF.

% 2. Describe the CERRA reanalysis data
% Explain the source dataset, temporal coverage, hourly construction, variables, and role in pre-training.

We use CERRA reanalysis data for model pre-training. CERRA covers Europe at 5.5~km horizontal resolution and includes both surface-level and pressure-level variables from 1984 onward. CERRA analyses are available every three hours; we construct a uniform hourly dataset by filling the intervening hours with the corresponding one- and two-hour forecasts. This gives the pre-training stage a long record from which the model can learn cloud evolution before adaptation to observations.

% 3. Describe the NWCSAF satellite observations
% Explain effective cloudiness, its physical meaning, resolution, and role as the observation-space target.

To adapt the CERRA-pretrained model to satellite observations, we use NWCSAF effective cloudiness for the initial cloud fields and prediction target during fine-tuning. NWCSAF is an EUMETSAT programme that supports operational nowcasting with satellite-derived cloud data from geostationary platforms. The effective-cloudiness product is derived from SEVIRI observations, mainly using the 10.8~$\mu$m infrared window channel together with cloud fraction and cloud emissivity \citep{Aminou1997,Kerdraon2021}. It ranges from 0 for clear sky to 1 for optically thick cloud, with intermediate values representing semi-transparent cloud layers. The effective-cloudiness fields have a spatial resolution of 3~km at nadir and are available every 15~min.

% 4. Describe the MEPS forecast data
% Explain the operational forecast system, its cloud-cover representation, and its role as dynamic forcing and baseline.

Satellite observations provide the cloud-cover target, but they do not provide the future atmospheric state required for forecasting. We therefore use forecasts from MEPS, a regional numerical weather prediction system based on HARMONIE-AROME cycle 43h2.2 \citep{Bengtsson2017}, operated by the meteorological institutes of Estonia, Finland, Norway, and Sweden. It is initialized every three hours and outputs forecasts at hourly lead times up to 66 hours ahead on a 2.5~km grid. In this study, atmospheric fields from MEPS serve as time-varying inputs to CloudCast v2, while MEPS total-cloud-cover forecasts serve as the NWP baseline.

% 5. Explain common preprocessing
% State that all products are reprojected or interpolated to the shared 5 km grid.

%Figure \ref{fig:tcc_comparison} compares CERRA, MEPS, and NWCSAF cloud fields on a shared grid. In this example, all three capture the same broad cloud pattern. Although CERRA and MEPS differ, both show less small-scale variability and fewer values near 0 and 1 than NWCSAF.

\subsection{Forecasting task}

% 1. Define the prediction target
% State that the task is hourly cloud-cover prediction on the common 5 km grid for lead times 1-12 h.
% combine with the next point to explain that the model produces one-hour forecasts that are rolled out autoregressively to produce the full forecast sequence.
% 3. Define the autoregressive rollout
% Explain how one-hour predictions are recycled as inputs to produce the full forecast sequence.

Each forecast case is initialized at time \(t\) over the entire common 5~km grid. The initial-state input consists of two gridded cloud-cover fields valid at \(t-1\) and \(t\). The model also receives static fields and atmospheric fields valid at the two history times and at each forecast lead. We refer to the time-varying inputs, including atmospheric variables, insolation, and temporal encodings, as dynamic forcings. The model predicts cloud-cover fields from \(t+1\) to \(t+12\). Depending on the training stage, these forecasts are generated either autoregressively or directly from the initial state. The atmospheric input fields must be available over the full forecast horizon; the model predicts only cloud cover, not the atmospheric variables themselves. During pre-training, CERRA supplies both the cloud-cover fields and dynamic forcings. During fine-tuning, NWCSAF supplies the cloud-cover history and prediction targets, while MEPS supplies the dynamic forcings; the same NWCSAF--MEPS input combination is used at inference.

% 2. Define the model inputs
% Describe the two initial cloud-cover fields, static fields, and future MEPS forcing fields.

The model inputs describe three aspects of the forecasting problem: the recent cloud field, the surrounding atmospheric state, and the spatial and temporal setting. The variables were selected based on their physical relevance to cloud evolution and prior modelling experience. Wind represents cloud transport, temperature and humidity describe the thermodynamic environment, and geopotential provides information about the large-scale atmospheric state. Insolation, temporal encodings, and static fields account for diurnal, seasonal, and geographical influences. We use atmospheric variables at 1000, 925, 850, 700, and 500 hPa to sample the lower and middle troposphere while keeping the number of input fields manageable. The variables are organized as anemoi-datasets \citep{Lang2024} and listed in Table \ref{tab:parameters}.

\begin{table}[ht]
\centering
\begin{tabular}{lll}
\toprule
Variable & Field type & Model role \\
\midrule
Cloud cover & Total-column field & Prognostic \\
U-component of wind & Pressure-level field & Dynamic forcing \\
V-component of wind & Pressure-level field & Dynamic forcing \\
Temperature & Pressure-level field & Dynamic forcing \\
Relative humidity & Pressure-level field & Dynamic forcing \\
Geopotential & Pressure-level field & Dynamic forcing \\
Insolation & Solar forcing field & Dynamic forcing \\
Julian day & Temporal encoding & Dynamic forcing \\
Local time & Temporal encoding & Dynamic forcing \\
Latitude & Coordinate field & Static forcing \\
Longitude & Coordinate field & Static forcing \\
Orography & Surface field & Static forcing \\
Land-sea mask & Surface field & Static forcing \\
\bottomrule
\end{tabular}
\caption{Variables used by the model. Pressure-level fields are included at 1000, 925, 850, 700, and 500 hPa; insolation and temporal encodings are precomputed using anemoi-datasets \citep{Lang2024}. Dynamic forcings vary over the forecast horizon, whereas static forcings are fixed in time.}
\label{tab:parameters}
\end{table}

\subsection{Model architecture}

% 1. Give the architecture overview
% Name the U-shaped shifted-window vision transformer and explain why local attention is needed.

\begin{figure}[h]
\centering
\includegraphics[width=0.95\textwidth]{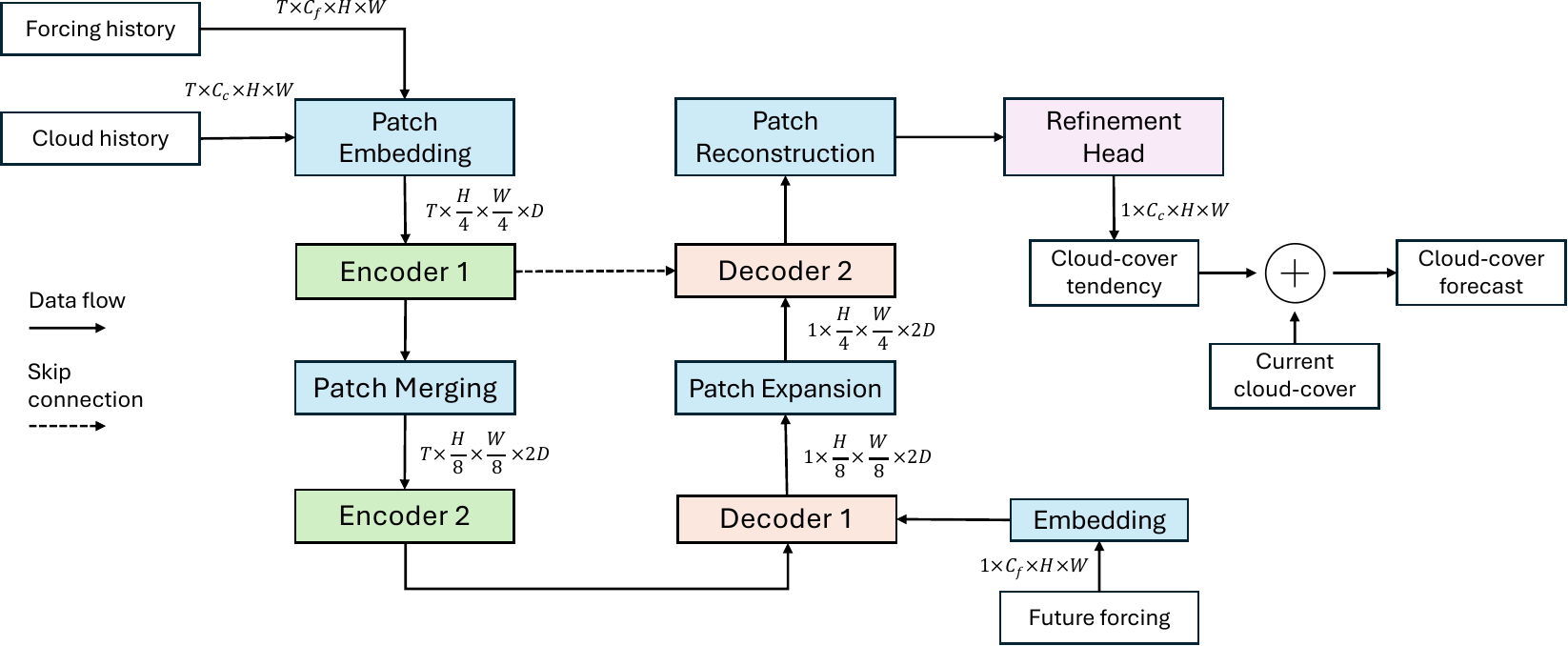}
\caption{Overview of the CloudCast v2 architecture for target lead time \(t+k\). Cloud-cover history and concurrent forcing history are embedded into patch tokens and processed by two encoder stages. The forcing fields valid at \(t+k\) are embedded separately and added before the first decoder stage. Patch expansion and reconstruction restore the spatial resolution, and the refinement head predicts a cloud-cover tendency, which is added to the current cloud cover to obtain the forecast. During direct prediction, the encoded history is shared and the decoder is applied independently for \(k=1,\ldots,12\). The dashed arrow denotes the encoder--decoder skip connection. Tensor shapes omit the batch dimension. Here, \(T=2\), \(C_c=1\), \(C_f=36\), \(D=256\), \(H=535\), and \(W=475\). The two additional conditioning channels used during flow-matching adaptation are not shown.}
\label{fig:cc2_architecture}
\end{figure}

Our model is a U-shaped vision transformer based on the Swin Transformer architecture \citep{Liu2021}. We represent the regional domain on a regular grid, allowing the model to treat the input fields as spatially ordered patches. Full self-attention \citep{Vaswani2017,Dosovitskiy2021} scales quadratically with the number of patches and would exceed the memory available for the regional domain. Swin instead computes attention within local windows and shifts the window partition between successive blocks. This reduces memory requirements while allowing information to propagate across window boundaries.

% 2. Describe the encoder-decoder structure
% Explain the two encoder blocks, two decoder blocks, patch merging, patch expansion, and prediction head.

The network follows an encoder–decoder structure with two encoder stages and two decoder stages, as shown in Figure \ref{fig:cc2_architecture}. he two-step cloud-cover and forcing histories are embedded into \(4\times4\)-pixel patches before entering the encoder. The two resolution levels combine local cloud structure with broader atmospheric context: on the 5~km grid, the attention windows span approximately 120~km in the first stage and 400~km in the second. The decoder combines information from both resolutions and reconstructs a full-resolution cloud-cover tendency, which is added to the current cloud field. The core architecture remains unchanged during pre-training and fine-tuning; only the training data and adaptation objective change.

% 3. Describe the block-level operations
% Explain attention, depth-wise convolution, and how the encoder and decoder blocks differ.

\begin{figure}[!htbp]
\centering
\includegraphics[width=0.95\textwidth]{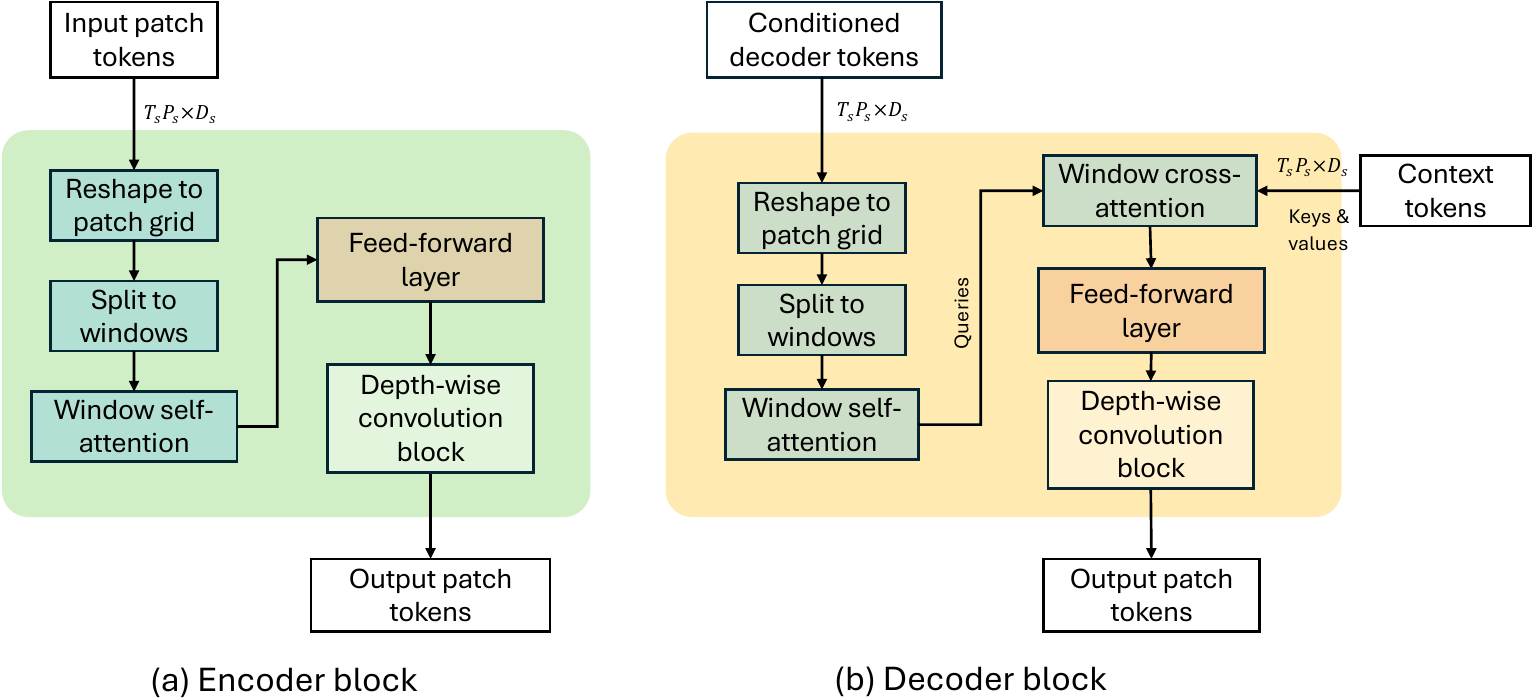}
\caption{Encoder and decoder block structure. (a) The encoder block arranges the embedded patches on their spatial grid, partitions the grid into local windows for self-attention, and then applies a feed-forward layer and depth-wise convolution for additional local spatial mixing. (b) The decoder block uses the same self-attention and refinement operations, but also applies window cross-attention to incorporate context tokens from the encoded representation or skip connection. Tensor shapes omit the batch dimension. At stage \(s\), the input, context, and output token sequences have shape \(T_sP_s\times D_s\), where \(P_s=H_sW_s\). The decoder tokens provide the queries in cross-attention, while the context tokens provide the keys and values. The decoder input has already been conditioned on the future forcing shown in Figure~\ref{fig:cc2_architecture}. Normalization and residual connections are omitted for clarity.}
\label{fig:encdec}
\end{figure}

Figure \ref{fig:encdec} summarizes the internal structure of the encoder and decoder blocks. Both combine windowed self-attention with depth-wise convolution, which adds local spatial mixing at relatively low computational cost. In the decoder, tokens conditioned on the forcing fields for the target lead time use cross-attention to incorporate context from the encoded representation and skip connection. During direct prediction, this allows the encoded initial state to be shared across lead times while each decoder output is conditioned on the corresponding atmospheric state.

% 4. Connect the text to the architecture figures
% Use the full architecture figure first, then the encoder-decoder block figure for implementation detail.

\subsection{Observation-space adaptation with conditional flow matching}
\label{sec:observation_space_adaptation}

% 1. Explain why adaptation is needed
% CERRA and NWCSAF represent cloud cover differently; naive fine-tuning loses forecast skill.

Long reanalysis records provide extensive training data for machine-learning weather models, but using satellite observations for initialization introduces a distribution shift. To transfer the CERRA-trained model to satellite observations, we first fine-tune it on NWCSAF effective cloudiness with a deterministic objective. This step adapts the model to observed cloud fields, but the resulting forecasts tend to lack fine-scale cloud structure: the adaptation remains difficult because the observational record is much shorter than the reanalysis record, and because CERRA total cloud cover and NWCSAF effective cloudiness represent related but different cloud quantities. To better preserve fine-scale cloud structure, we therefore add a generative adaptation stage based on conditional flow matching (CFM) \citep{Lipman2023}, with the clean-target parameterization described below.

% 2. Explain CFM conceptually
% CFM learns a vector field that moves samples from the simulation-trained distribution toward the observation-space target distribution.

Conditional flow matching is a generative modeling method that learns a transport from a simple source distribution to a target data distribution. Rather than producing only a deterministic point forecast, the model learns a conditional denoising process that can sample plausible fields from the target distribution. In our setting, the source distribution is Gaussian noise and the target is the observation-space distribution of cloud fields. During training, we sample a noise level \(\alpha \in [0,1]\), draw Gaussian noise \(z\), and construct the intermediate field in Equation~\eqref{eq:cfm_path}.

\begin{equation}
    \begin{aligned}[t]
        x_\alpha = (1-\alpha)y + \alpha z
    \end{aligned}
    \label{eq:cfm_path}
\end{equation}

where \(y\) is the NWCSAF effective-cloudiness target field. Unlike conventional CFM formulations, which train the model to predict the velocity along the noise-to-data path, we use a clean-target parameterization. This choice preserves the model's original cloud-cover prediction task, allowing initialization from the deterministic weights and continued use of the cloud-specific loss functions. The model receives \(x_\alpha\) and \(\alpha\), together with the encoded analysis fields and future forcings, and is trained to predict \(y\) directly. At inference, sampling starts from Gaussian noise at \(\alpha=1\). At each step, the model estimates the clean cloud field from the current intermediate field \(x_\alpha\) and uses this estimate to update \(x_\alpha\) at the next, lower value of \(\alpha\). Repeating this process over decreasing noise levels progressively transforms the initial noise into a cloud-cover forecast.

% 3. Explain how CFM is used here
% During adaptation, initialize from the pre-trained forecast model and optimize the CFM objective against NWCSAF targets.

We implement this clean-target formulation using the same CloudCast v2 architecture as the deterministic fine-tuned model. We initialize it from the deterministic weights, but we do not pass the deterministic mean forecast to the sampler. Instead, we provide \(x_\alpha\) and \(\alpha\) as additional future-forcing channels, while keeping the rest of the data encoding unchanged. This setup lets the model sample observation-space cloud fields while retaining the cloud-evolution dynamics learned during reanalysis pre-training.

\subsection{Training protocol}

% 1. Introduce the staged protocol
% State that training proceeds through CERRA pre-training, rollout stabilization, and NWCSAF adaptation.

\begin{figure}[h]
\centering
\includegraphics[width=1.0\textwidth]{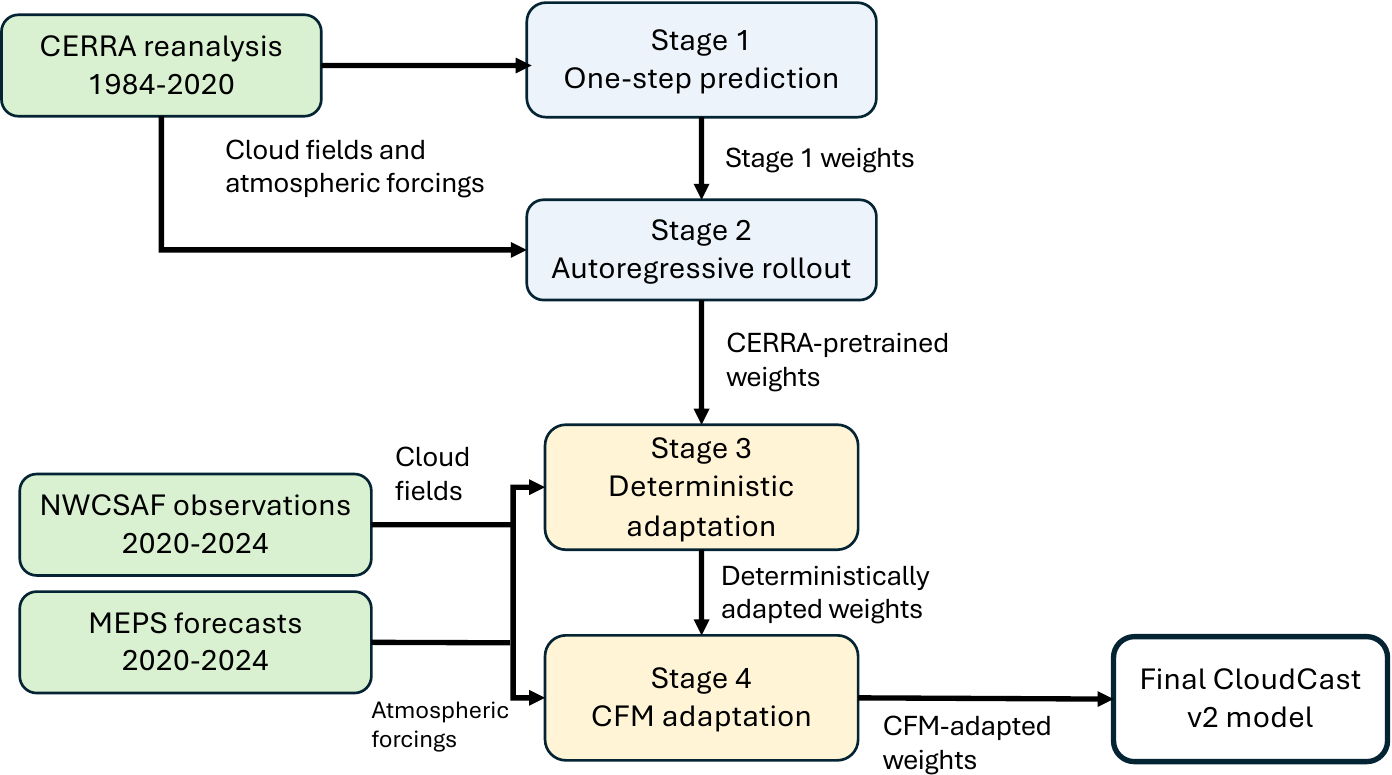}
\caption{Schematic overview of the four-stage CloudCast v2 training protocol. Stages 1--2 pre-train the model on CERRA reanalysis to learn cloud-evolution dynamics. Stages 3--4 adapt the model to NWCSAF effective cloudiness using MEPS forcings: stage 3 performs deterministic adaptation, and stage 4 applies conditional flow matching to produce the final observation-space forecast model. Arrow labels distinguish the training data supplied to each stage from the model weights passed forward through the training sequence.}
\label{fig:cc2_workflow}
\end{figure}

% 1. Describe stage 1 and stage 2
% Give the long CERRA pre-training setup: data, loss, rollout length, steps, and hardware.
% Explain scheduled sampling and rollout weighting for autoregressive stability.

We train the model in four sequential stages, moving from reanalysis pre-training to observation-space adaptation (Figure \ref{fig:cc2_workflow}). All stages are run on AMD GPUs on the LUMI supercomputer, and their configurations are summarized in Table \ref{tab:training}. 

The first two stages pre-train the model on CERRA. Stage 1 uses one-step prediction with mean squared error (MSE) loss for 100,000 iterations, giving the model an initial representation of cloud evolution from the full CERRA record. Stage 2 continues CERRA pre-training using six-hour autoregressive rollouts. At each step, the cloud fields supplied as history for the next prediction are taken either from CERRA or from the model's preceding forecasts. Scheduled sampling gradually increases the probability of using the model forecasts during training \citep{Bengio2015}. The loss initially emphasizes the first two forecast hours and gradually shifts toward equal weighting across all six hours. 

% 2. Describe fine-tuning stage 3 

In stage 3, we adapt the CERRA-trained model to satellite observations by fine-tuning it on NWCSAF effective cloudiness, which we treat as a bounded fractional target. We use direct prediction mode: the model predicts all 12 lead times from the same initial state instead of feeding its own predictions back autoregressively. Direct prediction avoids autoregressive error accumulation through the forecast sequence and lets each lead time use its corresponding MEPS forcing. The resulting model transfers the CERRA-trained dynamics to satellite-derived cloud fields and serves as the deterministic non-CFM ablation.

% 3. Describe stage 4
% Explain the concrete fine-tuning setup and refer back to the CFM adaptation subsection for the method.

Stage 4 applies CFM as a second observation-space adaptation stage, initialized from the deterministic stage-3 model. We train the model with the noisy intermediate fields described in Section \ref{sec:observation_space_adaptation}, while keeping the same direct prediction setup over the 12-hour horizon. This stage produces the final flow-matched CloudCast v2 model used in the main evaluation.

In both observation-space adaptation stages, we combine binary cross-entropy (BCE) loss with a differentiable approximation of the fractions skill score (FSS; \citealp{Roberts2008}) used in evaluation. We refer to this approximation as Soft-FSS. Soft-FSS replaces hard cloud-category thresholds with smooth thresholds and compares neighborhood cloud fractions over several spatial scales. We use the cloud-category thresholds from Table \ref{tab:categories} and neighborhood scales of 3, 6, and 12 grid cells.

% 5. Summarize hyperparameters in the table
% Use the table for learning rates, batch sizes, rollout lengths, losses, hardware, and flow settings.

\begin{table}[ht]
\centering
\begin{tabular}{lllll}
\toprule
Configuration & Stage 1 & Stage 2 & Stage 3 & Stage 4 \\
\midrule
Stage objective & CERRA one-step & CERRA rollout & NWCSAF deterministic & NWCSAF CFM \\
Optimizer & AdamW & AdamW & AdamW & AdamW \\
Learning rate & 5e-4 & 2e-5 & 3e-5 & 5e-6 \\
Scheduler & Cosine annealing & Cosine annealing & Cosine annealing & Cosine annealing \\
Steps & 100,000 & 20,000 & 20,000 & 20,000 \\
Warmup steps & 5,000 & 1,000 & 500 & 500 \\
Loss function & MSE & MSE & BCE + $0.2\times\text{Soft-FSS}$ & BCE + $0.2\times\text{Soft-FSS}$ \\
Rollout mode & One-step & Autoregressive & Direct & Direct \\
Rollout length & 1 & 6 & 12 & 12 \\
%Rollout weighting & No & Yes & No & No \\
Scheduled sampling & No & Yes & No & No \\
Effective batch size & 128 & 128 & 32 & 32 \\
Training data length & 1984-2020 & 1984-2020 & 2020-2024 & 2020-2024 \\
Training time (GPUh) & 2,200 & 2,300 & 480 & 750 \\
\bottomrule
\end{tabular}
\caption{Summary of the four-stage training protocol. Stages 1--2 pre-train the model on CERRA, stage 3 adapts it deterministically to NWCSAF effective cloudiness, and stage 4 applies CFM adaptation from the stage-3 checkpoint.}
\label{tab:training}
\end{table}

\subsection{Evaluation protocol and metrics}
\label{sec:evaluation_metrics}

% 0. Introduce the evaluation framework, including the reference models.

We evaluate forecasts on the fully held-out 2025 test year, with initializations every three hours. Verification covers lead times \(t+1\) to \(t+12\) against NWCSAF effective cloudiness on the common 5~km grid. After excluding initialization times with missing satellite scans or failed MEPS cycles, the common test set contains \(n=2{,}855\) initialization times.

We compare CloudCast v2 with two reference systems: CloudCast v1 \citep{Partio2025} and the MEPS control forecast \citep{Eresmaa2026}. We also include a deterministic CloudCast v2 ablation trained with the same NWCSAF fine-tuning setup but without the CFM stage; this isolates the contribution of CFM to forecast skill.  CloudCast v1 produces forecasts on its native \(512 \times 512\) grid, whereas CloudCast v2 uses the \(535 \times 475\) common verification grid. We therefore resample CloudCast v1 forecasts using nearest-neighbour interpolation. Consequently, CloudCast v1 has a small nonzero error at \(t+0\), despite using the same underlying observed initial field.

% 1. Introduce the metric families
% Explain why evaluation uses pixel-wise, spatial, and event-based metrics.

We evaluate forecast performance over the 1--12~h lead-time range from four complementary perspectives: (1) pixel-wise errors against NWCSAF effective cloudiness; (2) anomaly- and reference-based scores that allow comparison with MEPS despite cloud-variable differences; (3) neighborhood-based scores of spatial cloud placement; and (4) skill in cases with substantial cloud-cover changes. Because the reference systems predict partly different cloud quantities, we apply each metric only where the comparison is meaningful.

% 2. Define pixel-wise metrics
% Present bias and MAE and explain what each captures.

We begin with bias and mean absolute error (MAE), which apply most directly to the observation-space forecasts. Bias measures the mean signed error and indicates whether forecasts systematically overpredict or underpredict cloud cover, as defined by Equation~\eqref{eq:bias}.

\begin{equation}
    \begin{aligned}[t]
        \text{Bias} = \frac{1}{N} \sum_{i=1}^{N} (f_i - o_i)
    \end{aligned}
    \label{eq:bias}
\end{equation}

where \(f_i\) and \(o_i\) are the forecast and observed effective-cloudiness values, respectively, and \(N\) is the number of verified grid-point values. With this convention, positive values indicate overprediction and negative values indicate underprediction. MAE measures the mean error magnitude regardless of sign, as defined by Equation~\eqref{eq:mae}.

\begin{equation}
    \begin{aligned}[t]
        \text{MAE} = \frac{1}{N} \sum_{i=1}^{N} |f_i - o_i|
    \end{aligned}
    \label{eq:mae}
\end{equation}

We report bias and MAE only for systems that predict NWCSAF effective cloudiness directly. This includes CloudCast v1, CloudCast v2, the no-CFM ablation, and Eulerian persistence, which uses the initial observed cloud field as the forecast at every lead time. MEPS is excluded from bias and MAE because it forecasts cloud area fraction rather than effective cloudiness, creating a systematic offset in direct comparison. We include MEPS only in the ACC and MSESS analyses described below.

Anomaly- and reference-based skill scores provide a way to compare forecast systems beyond direct pointwise error. We use two metrics: the anomaly correlation coefficient (ACC) and the mean squared error skill score (MSESS) \citep{Murphy1989}. For each forecast system and the observations, we calculate a separate per-grid-point climatology for each valid calendar month and UTC hour by pooling all cases and lead times in the 2025 test set. ACC is then computed separately for each case and lead time using Equation~\eqref{eq:acc}.

\begin{equation}
\mathrm{ACC}
=
\frac{
    \sum_{i=1}^{N}\left(f_i - c_{i,m,h}^{(f)}\right)\left(o_{i} - c_{i,m,h}^{(o)}\right)
    }{
    \sqrt{\sum_{i=1}^{N}\left(f_i - c_{i,m,h}^{(f)}\right)^2}
    \sqrt{\sum_{i=1}^{N}\left(o_i - c_{i,m,h}^{(o)}\right)^2}
    }
\label{eq:acc}
\end{equation}

where \(f_i\) and \(o_i\) are the forecast and observed values at grid point \(i\), and \(c^{f}_{i,m,h}\) and \(c^{o}_{i,m,h}\) are their respective climatological means for valid month \(m\) and hour \(h\). The score evaluates spatial agreement between forecast and observed anomalies without removing their spatial means again. It ranges from \(-1\) to 1, with higher values indicating better anomaly-pattern agreement. Reported lead-time scores are averaged across cases.

MSESS compares the forecast MSE with that of Eulerian persistence, as defined by Equation~\eqref{eq:msess}. Both errors are calculated against NWCSAF effective cloudiness for all systems. The MEPS score therefore remains affected by the mismatch between cloud area fraction and effective cloudiness.

\begin{equation}
\mathrm{MSESS} = 1 - \frac{\text{MSE}_{fcst}}{\text{MSE}_{ref}}
\label{eq:msess}
\end{equation}

MSESS is 1 for a perfect forecast, 0 for skill equal to the reference forecast, and negative when the forecast is worse than the reference.

% 3. Define spatial verification
% Present FSS, neighborhood fractions, and why it is appropriate for cloud fields.

Pointwise and anomaly-based metrics do not account for small spatial displacement errors in coherent cloud structures. We therefore use FSS \citep{Roberts2008}, which evaluates spatial agreement by comparing the fraction of grid points exceeding a chosen threshold within a neighborhood around each point. The FSS is defined by Equation~\eqref{eq:fss}.

\begin{equation}
    \begin{aligned}[t]
        \text{FSS} = 1 - \frac{\sum_{i=1}^{N} (f_i - o_i)^2}{\sum_{i=1}^{N} f_i^2 + \sum_{i=1}^{N} o_i^2}
    \end{aligned}
    \label{eq:fss}
\end{equation}

where $f_i$ and $o_i$ are the forecast and observed fractions of grid points exceeding the threshold in the neighborhood around point $i$. The FSS ranges from 0 to 1, with higher values indicating better spatial agreement. In this study, FSS is computed over neighborhood scales from 25 to 110~km and reported separately for each cloudiness category, allowing us to assess whether the models place cloud structures correctly even when exact grid-point agreement is imperfect.

% 4. Define cloudiness thresholds
% Explain the cloud-cover categories used to threshold FSS and event detection.

Because MEPS predicts cloud area fraction rather than NWCSAF effective cloudiness, we exclude it from the category-based FSS comparison. Following \citet{Partio2025}, we divide effective cloudiness into four cloudiness categories (Table \ref{tab:categories}).

\begin{table}[ht]
\centering
\begin{tabular}{llll}
\toprule
Description & Cloudiness value & Octas & Grid-point fraction \\
\midrule
Clear sky & [0, 0.0625] & 0 & 0.24 \\
Partly cloudy & (0.0625, 0.5625] & 1-4 & 0.07 \\
Mostly cloudy & (0.5625, 0.9375] & 5-7 & 0.14 \\
Overcast & (0.9375, 1.0] & 8 & 0.55 \\
\bottomrule
\end{tabular}
\caption{Cloudiness categories used for threshold-based verification. The cloudiness value gives the NWCSAF effective-cloudiness interval defining each category. Octas give the corresponding traditional cloud-cover category in eighths, and grid-point fraction gives the proportion of grid points in the test dataset that fall into each category.}
\label{tab:categories}
\end{table}

% 5. Define genesis and lysis evaluation
% Describe how cloud formation and dissipation events are detected and scored.

Finally, we evaluate cases in which the observed cloud field changes substantially during the forecast window. For each forecast, we measure the domain-mean squared 12-hour change in observed effective cloudiness using Equation~\eqref{eq:dynamic_change}.

\begin{equation}
D = \frac{1}{N}\sum_{i=1}^{N}\left(o_{i,t+12} - o_{i,t}\right)^2
\label{eq:dynamic_change}
\end{equation}

where \(o_{i,t}\) is the observed effective cloudiness at grid point \(i\) and time \(t\). We rank all forecasts from the 2025 test year by \(D\) and retain the top 20\% as the dynamic subset, corresponding to 570 forecasts.

For these rapidly changing cases, horizontal displacement of the initial cloud field provides a more relevant baseline than simple persistence. We therefore construct a Lagrangian persistence baseline by advecting the initial cloud-cover field with Färneback optical flow \citep{Farnebck2003}, estimated from pre-forecast frames only, avoiding information from the verification period. Skill on the dynamic subset is reported as MSESS relative to this advected-persistence baseline, testing whether the model adds skill beyond displacement of the initial cloud field.

\section{Results}

% In this section, we evaluate the forecasts from four quantitative perspectives. First, we examine grid-point errors against NWCSAF effective cloudiness. Second, we assess anomaly and reference-based skill, which allows comparison with MEPS. Third, we evaluate neighborhood-scale spatial agreement across cloudiness regimes. Finally, we focus on cases with large observed cloud-cover changes to test whether the models add skill beyond advecting the initial cloud field. We then summarize the quantitative results and use case studies to illustrate the main trade-off between pixel-wise accuracy and neighborhood-scale spatial agreement.

This section compares CloudCast v2 with CloudCast v1, the CloudCast v2 no-CFM ablation, Eulerian persistence, and MEPS on the 2025 test set using the applicable metrics defined in Section~\ref{sec:evaluation_metrics}.

\subsection{Pixel-wise verification}

% Start by explaining the evaluation setup: test year, reference models, and metrics.

\begin{figure}[!htbp]
\centering
\includegraphics[width=1.0\textwidth]{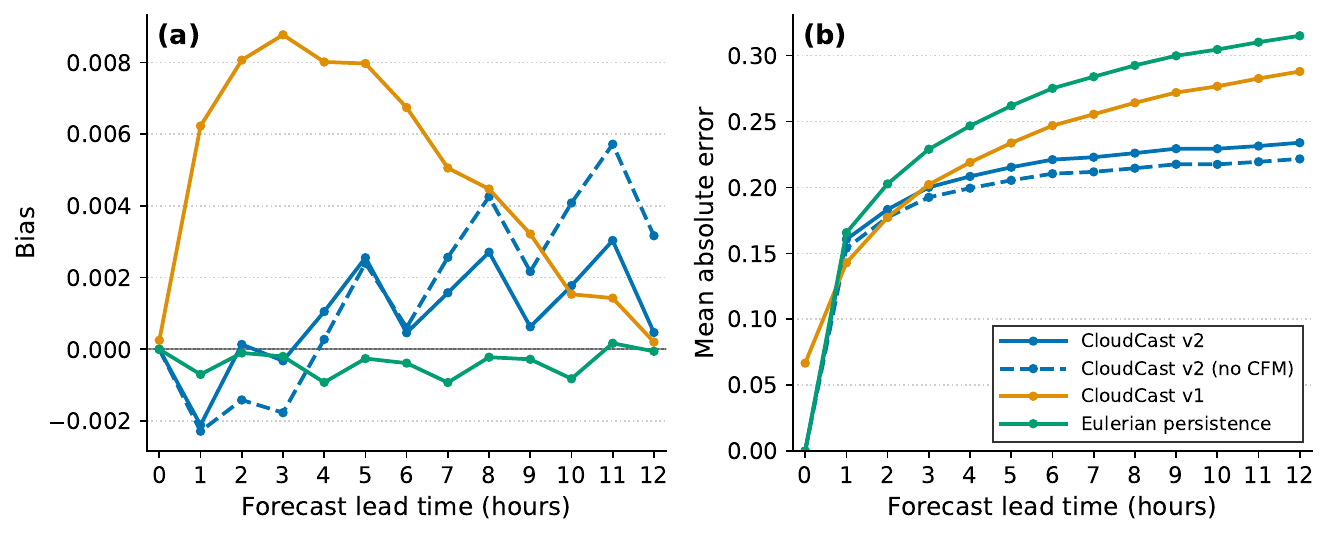}
\caption{Pixel-wise forecast errors relative to NWCSAF effective cloudiness over the 2025 test set. (a) Bias, with positive values indicating overprediction, and (b) MAE as a function of lead time. Curves show CloudCast v2, CloudCast v2 without CFM, CloudCast v1, and Eulerian persistence. Lead time \(t+0\) is included as an initialization check; the nonzero CloudCast v1 error reflects resampling from its native grid.}
\label{fig:mae}
\end{figure}

We first assess pointwise forecast errors using bias and MAE (Fig. \ref{fig:mae}). Panel (a) shows that both CloudCast v2 variants remain close to zero bias throughout the forecast window. CloudCast v1 has a positive bias that peaks near \(t+3\) and subsequently decreases. Because its MAE continues to increase over the same period, the decreasing bias indicates greater cancellation between positive and negative errors rather than improving pointwise accuracy.

Panel (b) shows that CloudCast v1 has the lowest MAE during the first two forecast hours. From approximately \(t+3\), both CloudCast v2 variants have lower MAE than CloudCast v1 and Eulerian persistence. Their MAE growth also slows after about \(t+5\), whereas the errors of CloudCast v1 and persistence continue to increase, widening the advantage of CloudCast v2 at longer lead times. The no-CFM model has consistently slightly lower MAE than the CFM model, indicating that the benefit of flow matching is not primarily expressed in pixel-wise absolute error.

\subsection{Pattern and reference-based skill}

% This section should answer: Does the model capture anomaly patterns and add value over persistence/MEPS despite cloud-cover definition differences?
% 
% Main message:
%    
% - ACC: pattern agreement.
% - MSESS: added value over reference.
% - This is where MEPS belongs more naturally than in MAE/bias.

\begin{figure}[!htbp]
\centering
\includegraphics[width=1.0\textwidth]{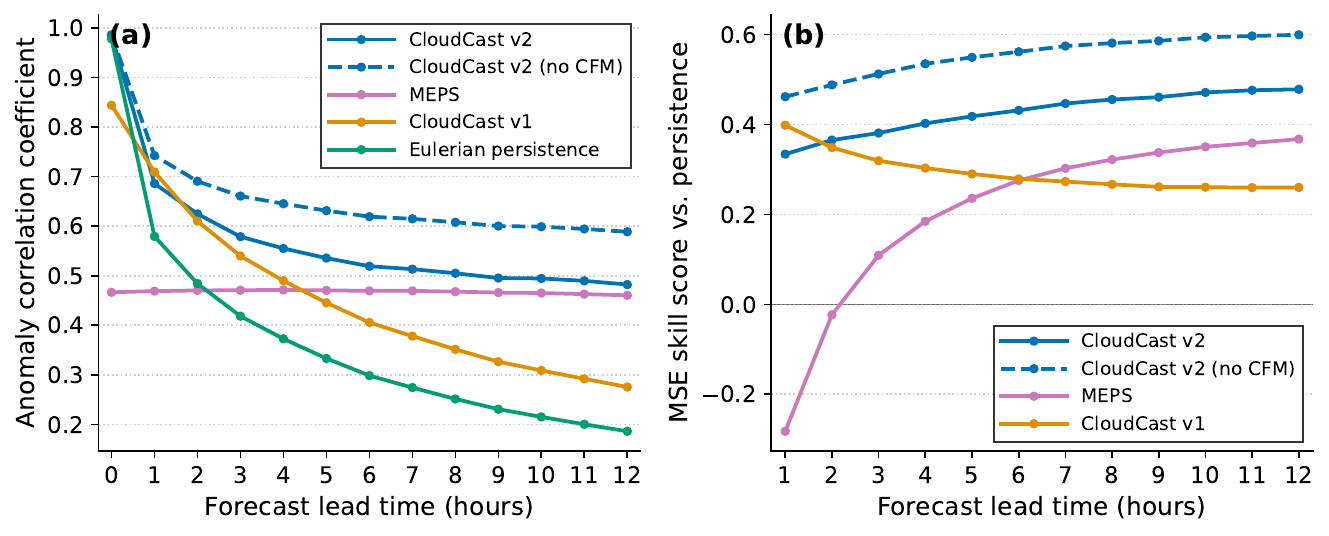}
\caption{Anomaly and reference-based skill on the 2025 test set: (a) ACC against NWCSAF effective cloudiness and (b) MSESS relative to Eulerian persistence. ACC includes \(t+0\) as an initialization check, whereas MSESS begins at \(t+1\) because it is undefined at \(t+0\). Panel (a) includes all five systems; in panel (b), the horizontal line at zero denotes equal skill to Eulerian persistence.}
\label{fig:acc_msess}
\end{figure}

Both CloudCast v2 variants maintain higher ACC than CloudCast v1, MEPS, and Eulerian persistence through most of the forecast window (Fig. \ref{fig:acc_msess}a). The no-CFM variant has the highest ACC at all lead times, while the full CFM model remains close to it and clearly above CloudCast v1 after the first few hours. CloudCast v1 falls below the nearly constant MEPS ACC at about \(t+5\), whereas both CloudCast v2 variants remain above MEPS through \(t+12\). Eulerian persistence decreases most rapidly as the initial observed cloud field becomes outdated. MEPS, in contrast, has nearly constant ACC with lead time. Because it is not initialized from the observed NWCSAF cloud field, its ACC is already relatively low at short lead times; the nearly flat curve indicates that this initial disagreement changes little over the forecast window.

MSESS is positive for both CloudCast v2 variants throughout the forecast window and increases with lead time as their advantage over persistence widens (Fig. \ref{fig:acc_msess}b). The no-CFM variant has the highest MSESS at every lead time; the full CFM model is below CloudCast v1 at \(t+1\) but exceeds both CloudCast v1 and MEPS thereafter. MEPS becomes skillful relative to persistence at \(t+3\) and overtakes CloudCast v1 at about \(t+7\), but remains below both CloudCast v2 variants through \(t+12\).

\subsection{Spatial skill across cloudiness regimes}

% This section should answer: Are cloud structures in roughly the right place and at the right scale?
% 
% Main message:
% 
% - FSS goes beyond pointwise matching.
% - Skill differs by cloudiness regime.
% - Intermediate categories are especially important because they are structurally complex.
% - This is where CFM may show value even if MAE does not.

\begin{figure}[!htbp]
\centering
\includegraphics[width=1.0\textwidth]{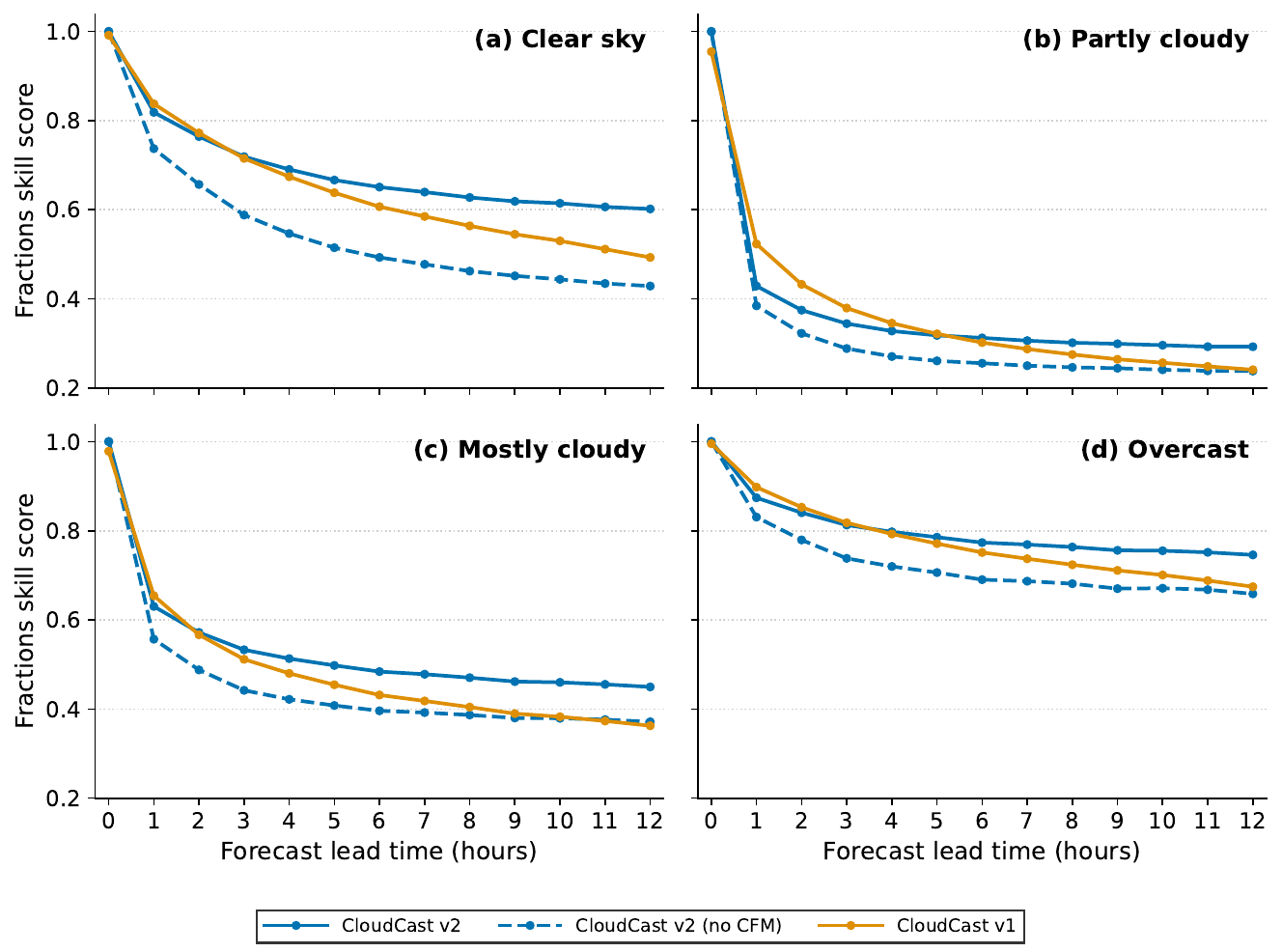}
\caption{FSS by cloudiness category as a function of forecast lead time, averaged over neighborhood scales from 25 to 110~km. Scores are computed against NWCSAF effective cloudiness for (a) clear sky, (b) partly cloudy, (c) mostly cloudy, and (d) overcast conditions, as defined in Table \ref{tab:categories}. All panels use the same y-axis scale. Curves show CloudCast v2 with and without CFM and CloudCast v1; higher values indicate better neighborhood-scale spatial agreement.}
\label{fig:fss}
\end{figure}

To assess neighborhood-scale spatial agreement, Figure \ref{fig:fss} shows FSS across the four cloudiness categories defined in Table \ref{tab:categories}, averaged over neighborhood scales from 25 to 110~km.

Across the forecast window, FSS remains substantially higher for clear-sky and overcast conditions than for the two intermediate categories. The lower skill for partly and mostly cloudy conditions is consistent with the greater spatial complexity of intermediate cloud fields.

CloudCast v1 performs best at \(t+1\) in every category, but its skill declines more rapidly. The full CloudCast v2 overtakes it at approximately \(t+2\) for mostly cloudy, \(t+4\) for clear-sky and overcast, and \(t+6\) for partly cloudy conditions, remaining more skillful thereafter. In contrast, the no-CFM variant remains below CloudCast v1 over most of the forecast window and only matches or exceeds it at longer lead times for mostly cloudy conditions. The full model achieves higher FSS than the no-CFM variant across nearly all lead times and categories, with the difference increasing most clearly with lead time under clear-sky conditions. These comparisons indicate that the CFM stage accounts for most of CloudCast v2's long-lead FSS advantage over CloudCast v1.

\subsection{Skill during large cloud-cover changes}

\begin{figure}[!htbp]
\centering
\includegraphics[width=1.0\textwidth]{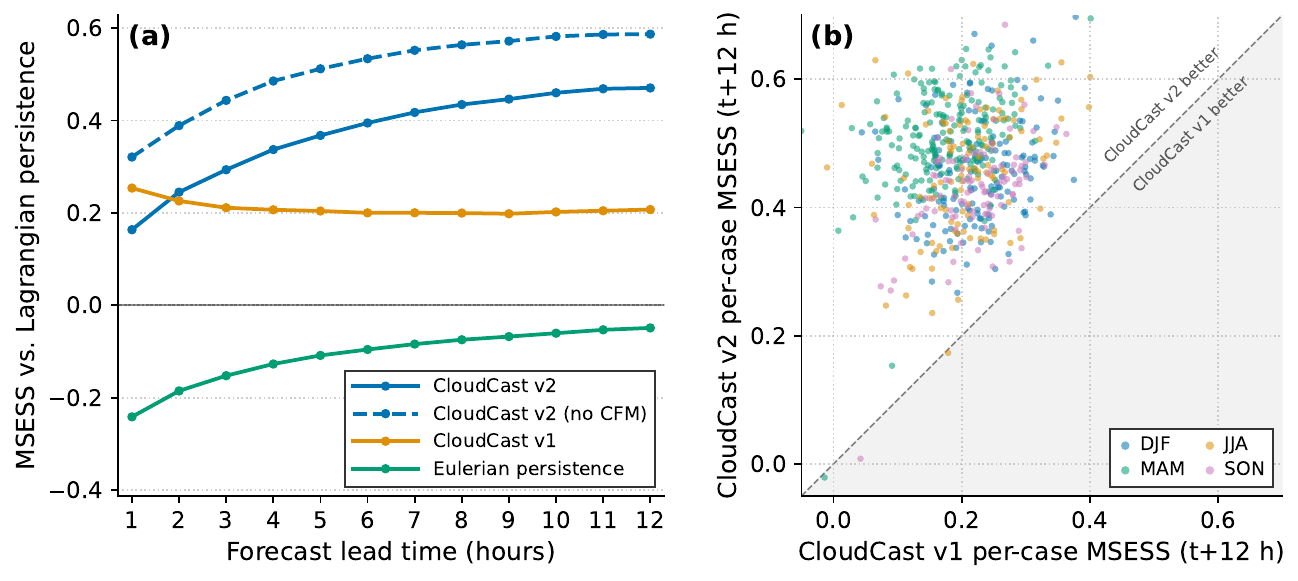}
\caption{Skill on the top-quintile dynamic cases, evaluated with MSESS relative to Lagrangian persistence. (a) Mean MSESS as a function of forecast lead time. (b) Per-case MSESS at \(t+12\) for CloudCast v2 and CloudCast v1, with points colored by season. Points above the diagonal indicate higher skill for CloudCast v2.}\label{fig:genesis_lysis}
\end{figure}

Figure \ref{fig:genesis_lysis} shows forecast skill on the top-quintile dynamic subset defined in Section \ref{sec:evaluation_metrics}, using Lagrangian persistence as the reference.

% This section should answer: Can the model predict cloud changes that are not explained by transport alone?
%
% Main message:
%
% - You select high-change cases.
% - Advected persistence is the reference.
% - Positive MSESS means the model adds skill beyond moving the initial cloud field.
% - This section should be careful: it does not directly “detect” genesis/lysis, but tests skill in cases where formation/dissipation likely matter.

%Both CloudCast v2 variants add skill beyond advecting the initial observed cloud field throughout the 12-hour forecast window (Fig. \ref{fig:genesis_lysis}a). CloudCast v1 also remains skillful relative to Lagrangian persistence, but its skill is lower and nearly flat with lead time. Eulerian persistence is worse than the Lagrangian reference, suggesting that horizontal displacement explains part of the observed cloud-cover change.

In the lead-time comparison (panel a), both CloudCast v2 variants add skill beyond advecting the initial observed cloud field throughout the 12-hour forecast window. The no-CFM variant has the highest MSESS at every lead time, while the full model remains clearly above CloudCast v1. The relative skill of both variants increases with lead time as Lagrangian persistence becomes a weaker forecast. CloudCast v1 remains skillful but nearly constant, whereas Eulerian persistence performs worse than Lagrangian persistence throughout the forecast window.

%The case-by-case comparison shows that CloudCast v2 improves on CloudCast v1 for most high-change cases at \(t+12\) (Fig. \ref{fig:genesis_lysis}b). The improvement is visible across seasons, with most points lying above the diagonal in each season. The gains appear largest in spring and smaller in autumn, while summer shows greater case-to-case variability. This seasonal structure may reflect stronger diurnally driven cloud development in spring and more persistent cloudy conditions in autumn. Because \(t+12\) lies beyond CloudCast v1's five-hour training horizon, we also performed the per-case comparison at \(t+5\). CloudCast v2 achieved higher MSESS in 96.3\% of the dynamic cases (not shown), indicating that its advantage is not confined to lead times outside the CloudCast v1 training range. These results suggest that CloudCast v2 better predicts cloud evolution beyond horizontal transport, especially when formation or dissipation likely contributes to the observed cloud-cover change.

In the case-by-case comparison at \(t+12\) (panel b), CloudCast v2 outperforms CloudCast v1 for most high-change cases, with most points lying above the diagonal in every season. CloudCast v2 also retains positive MSESS in nearly all cases, including many in which CloudCast v1 has little or negative skill. Because \(t+12\) lies beyond CloudCast v1's five-hour training horizon, we also performed the comparison at \(t+5\). CloudCast v2 achieved higher MSESS in 96.3\% of the dynamic cases (not shown), indicating that its advantage is not confined to lead times outside the CloudCast v1 training range.

\subsection{Overall model performance}

The verification metrics capture different aspects of forecast quality, and no single model performs best across all scores. Table \ref{tab:main_results} summarizes the quantitative results.

% Main results table — CFM cloud-cover paper (main-text version, with FSS column group).
% Disclosures to carry in Methods prose (formerly footnotes):
%  (1) MEPS: MAE/Bias and FSS omitted due to the +0.079 cloud-definition offset vs NWCSAF
%      (FSS category occupancy distorted by the offset); ACC is offset-invariant. MSESS is
%      reported, but the offset inflates MEPS MSE (~+offset^2), so its MSESS is a conservative
%      lower bound on MEPS skill; the strongly negative t+1 value reflects persistence being
%      near-perfect at short lead rather than a MEPS failure.
%  (2) CCv1: nearest-neighbour regrid from native 512x512; misregistration inflates MAE by
%      <=0.009 and reduces correlation by <=0.023; FSS unaffected; own-truth sensitivity in appendix.
%  (3) Population: n=2849 common inits, full-year 2025, 3-hourly; 6 frames from two zero-filled
%      missing scans excluded for all models; t+0 excluded from averages; eval year fully held out
%      (train end 2024-01-01, val end 2025-01-01).
%  (4) Lag-MSESS: top-quintile-dynamic subset (n=573, D=MSE(truth_t12, truth_t0)); baseline =
%      Farneback optical-flow advection of the observed t+0 field (pre-forecast frames only);
%      CIs = block bootstrap (block 12, B=5000). CFM-vs-CCv1 paired diff at t+12:
%      +0.267 [+0.248, +0.286], excludes zero.
% Bolding rule: best among FORECAST systems per column
% FSS partly-cloudy: CFM and CCv1 within the noise floor -> both bold (tie).
% Requires: booktabs.
\begin{table*}[!htbp]
\centering
\caption{Summary of verification metrics on the 2025 test set. MAE, bias, and FSS are averaged over \(t+1\)--\(t+12\), and FSS is additionally averaged over neighborhood scales from 25 to 110~km. Dynamic MSESS is computed on the top-quintile dynamic subset relative to Lagrangian persistence. Arrows indicate whether lower or higher values are better; for signed bias, values closest to zero are best. Bold values mark the best-performing forecast system or systems in each column.}
\label{tab:main_results}
\setlength{\tabcolsep}{3pt}
\scriptsize
\begin{tabular}{lrrrrrrrrrrrr}
\toprule
 & \multicolumn{2}{c}{Pixel-wise} &
   \multicolumn{2}{c}{ACC} &
   \multicolumn{2}{c}{MSESS} &
   \multicolumn{2}{c}{Dynamic MSESS} &
   \multicolumn{4}{c}{FSS} \\
\cmidrule(lr){2-3}\cmidrule(lr){4-5}\cmidrule(lr){6-7}\cmidrule(lr){8-9}\cmidrule(lr){10-13}
Model &
  MAE$\downarrow$ & Bias &
  $t{+}1\uparrow$ & $t{+}12\uparrow$ &
  $t{+}1\uparrow$ & $t{+}12\uparrow$ &
  $t{+}4\uparrow$ & $t{+}12\uparrow$ &
  Clear$\uparrow$ & Partly$\uparrow$ & Mostly$\uparrow$ & Overc.$\uparrow$ \\
\midrule
CloudCast\,v2 & 0.213 & $\mathbf{+0.001}$ & 0.686 & 0.483 & 0.335 & 0.479 &
  $+0.336$ & $+0.471$ &
  \textbf{0.668} & \textbf{0.325} & \textbf{0.500} & \textbf{0.785} \\
CloudCast\,v2 (no CFM) & \textbf{0.203} & $+0.002$ & \textbf{0.743} & \textbf{0.589} & \textbf{0.462} & \textbf{0.600} & $\mathbf{+0.485}$ & $\mathbf{+0.588}$ &
  0.520 & 0.270 & 0.416 & 0.708 \\
CloudCast\,v1 & 0.238 & $+0.005$ & 0.710 & 0.276 & 0.398 & 0.260 & $+0.202$ & $+0.205$ & 0.623 & 0.323 & 0.452 & 0.760 \\
MEPS (NWP) & --- & --- & 0.469 & 0.461 & $-0.282$ & 0.368 & --- & --- & --- & --- & --- & --- \\
% Eulerian persistence & 0.265 & $-0.000$ & 0.580 & 0.186 & 0.000 & 0.000 & $-0.128$ & $-0.050$ & 0.701 & 0.392 & 0.528 & 0.841 \\
\bottomrule
\end{tabular}
\end{table*}

The ablation comparison isolates the effect of CFM. The no-CFM model gives the lowest MAE and highest ACC and MSESS, whereas the full model gives higher FSS across all cloudiness categories. Thus, CFM slightly reduces performance according to conventional error-based metrics while improving neighborhood-scale spatial agreement.

Relative to CloudCast v1, both CloudCast v2 variants become more advantageous as lead time increases and perform better in cases with large observed cloud-cover changes. CloudCast v1 remains competitive at the earliest lead times, but its skill deteriorates more rapidly. CloudCast v2 therefore provides more sustained skill across the 12-hour forecast window, including when the observed cloud field changes substantially.

\subsection{Illustrative case studies}

% This should show the reader what the metrics look like in an actual forecast.
%
% Main message:
%
% - Choose one case that illustrates the quantitative findings.
% - Show where CloudCast v1/persistence fails, where no-CFM differs, and what CFM changes.
% - Do not make the case study carry the proof; it should make the earlier results tangible.

In addition to the quantitative evaluation, we examine two forecast cases to make the main model differences visible. The first case illustrates how CloudCast v2 captures regional cloud-cover evolution, and how CFM can sharpen cloud-edge structure. The second case shows a failure mode in which the same sharpening produces a confident false cloud structure.

\begin{figure}[!htbp]
\centering
\includegraphics[width=0.95\textwidth]{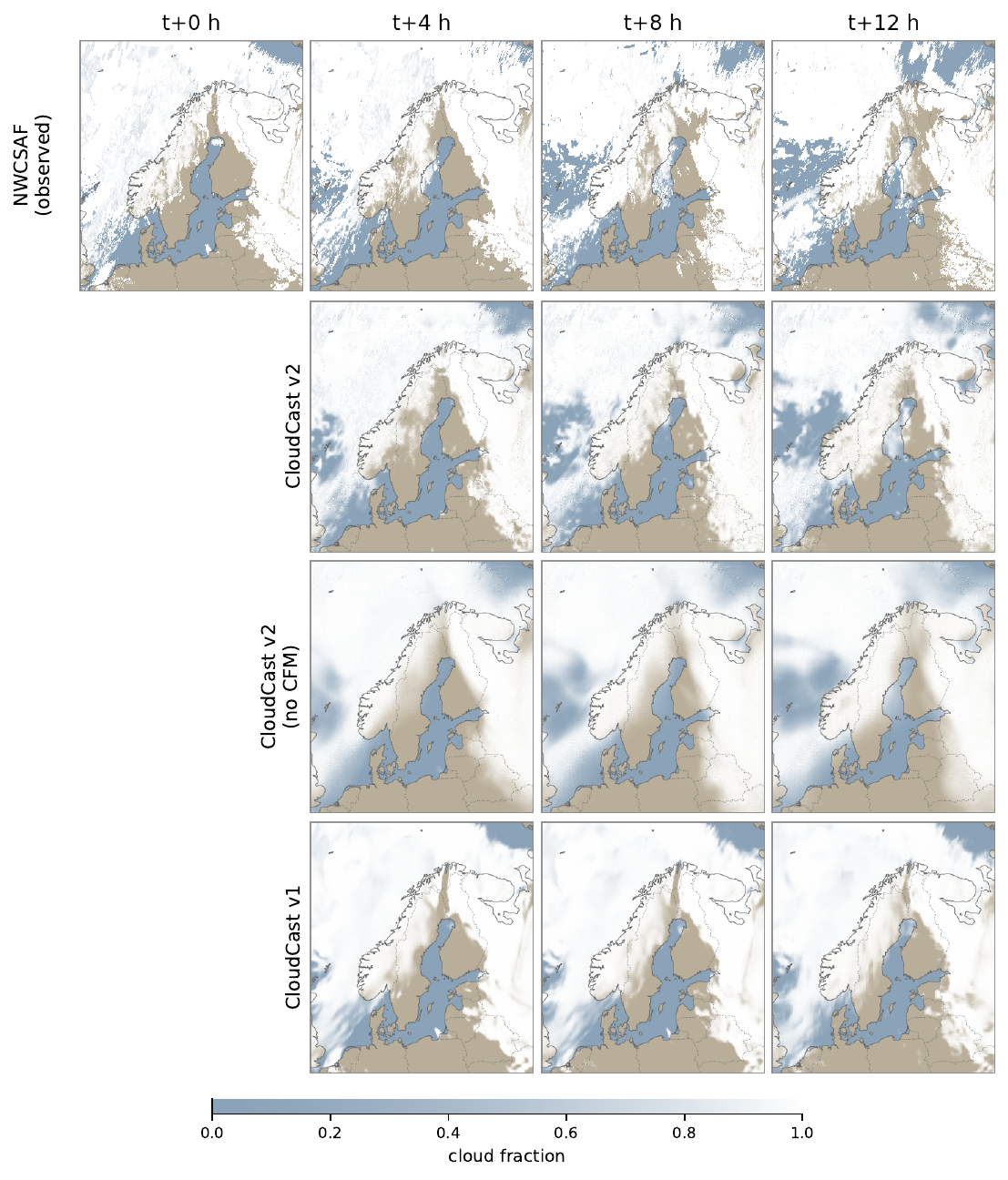}
\caption{Illustrative cloud-cover case initialized on 1 July 2025 at 03 UTC. Columns show the observed initial state and forecast lead times \(t+4\), \(t+8\), and \(t+12\) h; rows show NWCSAF observations and forecasts from CloudCast v2, CloudCast v2 without CFM, and CloudCast v1. Forecast rows begin at \(t+4\) because all models share the observed \(t+0\) initial field. The case has a near-neutral domain-mean cloud-cover change, but opposing regional signals: clearing over western Scandinavia and increasing cloud cover near the western Finnish coast.}
\label{fig:case_20250701_03Z}
\end{figure}

Figure \ref{fig:case_20250701_03Z} shows an extensive cloud field over western and northern Scandinavia at initialization, with comparatively clear conditions over Finland and the Baltic region. Over the following 12 hours, the domain-mean cloud cover changes little, but this masks opposing regional changes: clearing over western Scandinavia and increasing cloud cover near the western Finnish coast. CloudCast v1 fails to capture both signals, retaining too much cloud over western Scandinavia and missing the cloud increase near Finland. Both CloudCast v2 variants capture the broader evolution, but the no-CFM version produces smoother cloud edges and underestimates the spatial extent of the clearing. The full CFM model better preserves the sharp cloud-edge structure in both regions, even when pixel-wise errors are similar.

\begin{figure}[!htbp]
\centering
\includegraphics[width=0.95\textwidth]{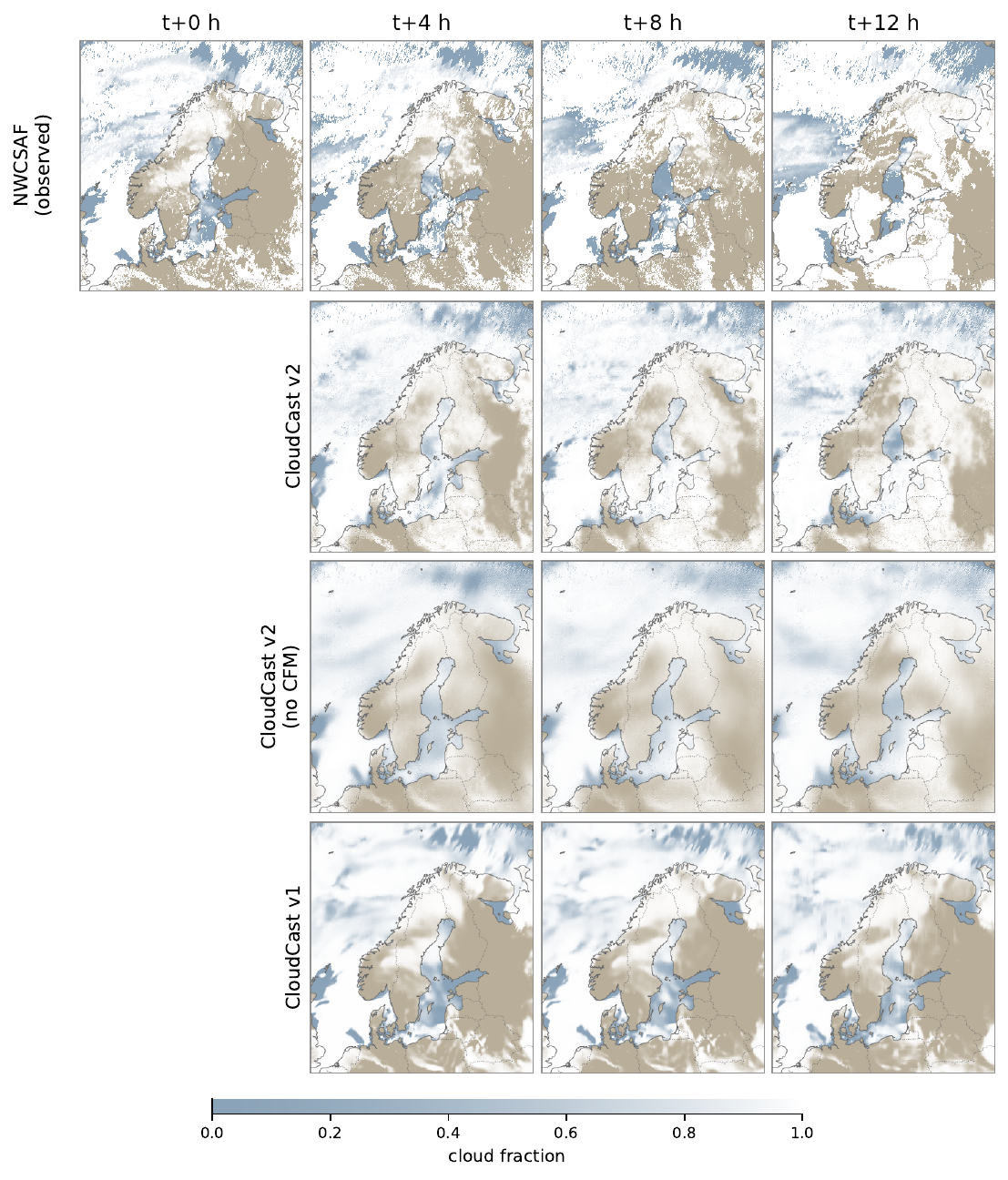}
\caption{Illustrative winter cloud-cover case initialized on 18 January 2025 at 21 UTC, showing winter cloud-cover evolution over Scandinavia. Columns show the observed initial state and forecast lead times \(t+4\) h, \(t+8\) h, and \(t+12\) h; forecast rows are shown only for the lead times because the models use the observed \(t+0\) cloud field as the initial state. Rows compare NWCSAF observations with CloudCast v2, CloudCast v2 without CFM, and CloudCast v1. The case shows a premature cloud structure produced by the CFM model over the southern part of the domain.}
\label{fig:failure_case_20250118_21Z}
\end{figure}

Figure \ref{fig:failure_case_20250118_21Z} begins with fragmented cloud cover over western Scandinavia and along the southern part of the domain, while much of Finland and the Baltic region is comparatively clear. Over the following 12 hours, the observed cloud cover expands across the southern and eastern parts of the domain. CloudCast v2 anticipates this development too early, producing a coherent cloud mass at \(t+4\) h and \(t+8\) h that is not yet present in the NWCSAF observations. By \(t+12\) h, the forecast more closely resembles the observed cloudier pattern, suggesting that the error is partly a timing error rather than a wholly spurious structure. The no-CFM version avoids the early false positive but remains smoother, while CloudCast v1 misses much of the subsequent cloud development over the Baltic countries. This case shows that the sharper structures introduced by CFM can occur at the wrong time.

\FloatBarrier
\section{Discussion and Conclusions}

% Start by stating the main result in plain terms.
% The model can produce a single 12-hour forecast sequence from observed initial conditions.
% It improves over CloudCast v1 especially at longer lead times and in structure-aware metrics.
% Avoid listing all metrics again; summarize the pattern.

In this study, we develop CloudCast v2, a machine-learning model that produces cloud-cover forecasts up to 12 hours from satellite-derived initial conditions and NWP atmospheric forcings. The model first learns cloud-evolution dynamics from a long reanalysis record and is then adapted to satellite-derived effective cloudiness. CloudCast v1, the earlier U-Net-based model, remains competitive at the earliest lead times. CloudCast v2, however, maintains higher skill later in the forecast window, achieves greater neighborhood-scale spatial agreement, and performs better in cases with large observed cloud-cover changes.

% Connect the result back to the nowcasting-to-forecasting seam.
% Emphasize that short-range forecasts benefit from observed cloud placement,
% but longer lead times require learned/dynamical evolution.
% CloudCast v2 is useful because it starts from observations but has learned evolution from reanalysis.

CloudCast v2 connects observation-based nowcasting with NWP-informed forecasting. Satellite fields anchor the forecast to the observed cloud distribution at initialization, while reanalysis pre-training provides a long record from which to learn cloud evolution. NWP atmospheric forcings then describe how the surrounding atmospheric conditions develop over the forecast period. This combination allows the cloud field to evolve beyond persistence or advection without losing its connection to the observed initial state. Its benefit becomes most apparent after the first few hours, as the skill of persistence and CloudCast v1 declines.

% Explain why no-CFM can look better in MAE/ACC while CFM looks better in FSS.
% Deterministic regression produces smoother fields and is favored by pixel-wise losses.
% CFM sharpens and structures the forecast, improving neighborhood-scale realism,
% but can increase local errors when features are displaced or falsely introduced.
% Link this to the positive and limitation case studies.

The benefit of CFM is not uniform across all metrics. Deterministic regression models tend toward conditional-mean forecasts, which are favored by pixel-wise losses such as MSE. The flow-matching stage produces sharper and more spatially coherent cloud structures, as reflected in its higher FSS, but it can also increase local errors when sharpened features are displaced or timed incorrectly. This trade-off is important operationally: sharper forecasts may better represent cloud-field morphology, but they can also produce confident local false positives or false negatives.

% Discuss the Lagrangian-persistence comparison.
% Positive skill relative to advected persistence suggests the model is doing more than moving the initial cloud field.
% Be careful: this does not prove explicit cloud genesis/lysis detection.
% It shows skill in cases where cloud formation, dissipation, or deformation are likely important.

The dynamic-case evaluation provides a complementary view of forecast skill. In the first forecast hours, much of the skill comes from starting with the observed cloud field, but longer lead times require the model to predict changes that are not explained by horizontal transport alone. We therefore compared the models against a Lagrangian persistence baseline that advects the initial cloud field using optical flow. Positive skill relative to this baseline shows that CloudCast v2 improves on simply displacing the initial cloud field. This does not prove that the model explicitly represents cloud formation or dissipation; rather, it indicates skill in cases where formation, dissipation, or deformation likely contribute to the observed cloud-cover change.

% Explain what the MEPS comparison does and does not show.
% MEPS predicts cloud area fraction, not NWCSAF effective cloudiness.
% Therefore direct pointwise comparison is not fully fair.
% ACC is still useful because it asks whether the large-scale anomaly pattern is placed correctly.
% Avoid claiming that CloudCast v2 is generally better than NWP in all senses.

The comparison with MEPS, used here as the NWP baseline, requires caution: MEPS forecasts cloud area fraction, whereas CloudCast v2 forecasts NWCSAF effective cloudiness, which also serves as the verification field. ACC reduces sensitivity to the resulting systematic offset by evaluating anomalies after separate month-and-hour climatologies are removed from each system and the observations. CloudCast v2 maintains higher ACC than MEPS throughout the 12-hour forecast window, indicating closer agreement with the observed large-scale anomaly pattern. MSESS also favors CloudCast v2, but because it is computed against NWCSAF effective cloudiness, the comparison remains affected by the difference between the predicted cloud quantities. MEPS nevertheless becomes more competitive as lead time increases, consistent with the diminishing value of the observed initial cloud field.

% Mention operational implications

MEPS also has an operational role in CloudCast v2 by supplying its atmospheric forcings. The latest NWCSAF fields provide the initial cloud state, and once both inputs are available, CloudCast v2 generates the complete 12-hour forecast in approximately 90 seconds on an NVIDIA H100 GPU. This short inference time adds little computational delay before forecast dissemination compared with producing the underlying NWP forecast. CloudCast v2 cannot be issued before the required MEPS fields become available, however, and therefore does not replace the NWP production cycle. Instead, it rapidly combines available NWP guidance with the latest satellite-observed cloud field.

% Mention the main limitations:
% - verification depends on NWCSAF effective cloudiness;
% - CFM can introduce confident false structures;
% - case studies are selected and illustrative;
% - evaluation is regional and one-year;
% - MEPS comparison is limited by variable-definition differences.
% Keep this paragraph compact.

Several limitations should be considered when interpreting these results. First, the verification uses NWCSAF effective cloudiness as the reference field. This provides a consistent observation-space target, but it does not capture all aspects of cloud cover that may matter for downstream applications such as radiation or solar-power forecasting. Second, the model is trained and evaluated on a single regional domain, and the evaluation covers only one held-out year. The model therefore cannot be assumed to transfer directly to other regions without domain-specific adaptation; similarly, the test set may not represent the full range of weather regimes encountered in other years. 

% Future work should follow from limitations:
% - probabilistic or ensemble CFM forecasts;
% - calibration of sharp CFM outputs;
% - extension to more regions/seasons;
% - better coupling with NWP predictors;
% - evaluating user-relevant quantities such as solar irradiance.

Future work should extend both the forecasting framework and the evaluation. Because the CFM formulation is stochastic, it could be developed into probabilistic or ensemble forecasts that quantify uncertainty in cloud evolution. Evaluation over additional years and geographic domains is also needed to assess the model's robustness and transferability. The direct-prediction setup could also support sub-hourly forecasts, provided that suitable forcing fields are available at the target times. Finally, the atmospheric forcing set was not extensively optimized in this study; a broader or more carefully selected set of predictors may further improve forecast quality. Together, these directions would test whether the same observation-initialized framework can support more flexible and user-relevant cloud forecasts.

% Final paragraph should not introduce new caveats.
% Restate the contribution: a unified observation-initialized 12-hour cloud-cover forecast model.
% Emphasize the practical value of bridging nowcasting and forecasting.

The results indicate that machine-learning models initialized from satellite observations can extend cloud-cover prediction beyond short-range extrapolation when supplied with NWP atmospheric fields. CloudCast v2 complements rather than replaces NWP by combining its atmospheric forecasts with the latest observed cloud field at a small additional computational cost.

\section*{Acknowledgements}

We acknowledge CSC – IT Center for Science, Finland for awarding this project access to the LUMI supercomputer, owned by the EuroHPC Joint Undertaking, hosted by CSC (Finland) and the LUMI consortium through the Finnish LUMI Extreme Scale call.
\\
We acknowledge the computational resources provided by the Aalto Science-IT project.

\section*{Additional information}

\textbf{Competing interests:} The authors declare no competing interests.

\section*{Data Availability Statement}

The source code used in this study is available at \url{https://github.com/fmidev/cloudcast}. CERRA data are publicly available from the Copernicus Climate Data Store at \url{https://doi.org/10.24381/cds.622a565a}. MEPS forecast data are openly available through the MET Norway THREDDS archive at \url{https://thredds.met.no/thredds/catalog/metno.html}. The NWCSAF effective-cloudiness data and the processed training and evaluation datasets are not publicly archived because of their volume and associated storage and distribution requirements.

\bibliographystyle{apalike}
\bibliography{references}

\end{document}